\pdfoutput=1 %

\documentclass[11pt,a4paper]{article}
\usepackage[hyperref]{acl2019}
\usepackage{times}
\usepackage{latexsym}

\usepackage{url}

\usepackage[utf8]{inputenc}
\usepackage[T1]{fontenc} %
\usepackage{multirow}
\usepackage{booktabs}
\usepackage{amsmath}
\usepackage{amsfonts}

\usepackage{seqsplit}

\aclfinalcopy %

\title{Bilingual Lexicon Induction through Unsupervised Machine Translation}

\author{Mikel Artetxe, Gorka Labaka, Eneko Agirre \\
  IXA NLP Group \\
  University of the Basque Country (UPV/EHU) \\
  \texttt{\{mikel.artetxe, gorka.labaka, e.agirre\}@ehu.eus}}

\date{}

\begin{document}
\maketitle
\begin{abstract}
A recent research line has obtained strong results on bilingual lexicon induction by aligning independently trained word embeddings in two languages and using the resulting cross-lingual embeddings to induce word translation pairs through nearest neighbor or related retrieval methods. In this paper, we propose an alternative approach to this problem that builds on the recent work on unsupervised machine translation. This way, instead of directly inducing a bilingual lexicon from cross-lingual embeddings, we use them to build a phrase-table, combine it with a language model, and use the resulting machine translation system to generate a synthetic parallel corpus, from which we extract the bilingual lexicon using statistical word alignment techniques. As such, our method can work with any word embedding and cross-lingual mapping technique, and it does not require any additional resource besides the monolingual corpus used to train the embeddings. When evaluated on the exact same cross-lingual embeddings, our proposed method obtains an average improvement of 6 accuracy points over nearest neighbor and 4 points over CSLS retrieval, establishing a new state-of-the-art in the standard MUSE dataset.
\end{abstract}

\section{Introduction}
\label{sec:introduction}

Cross-lingual word embedding mappings have attracted a lot of attention in recent times. These methods work by independently training word embeddings in different languages, and mapping them to a shared space through linear transformations. While early methods required a training dictionary to find the initial alignment \citep{mikolov2013exploiting}, fully unsupervised methods have managed to obtain comparable results based on either adversarial training \citep{conneau2018word} or self-learning \citep{artetxe2018robust}.

A prominent application of these methods is Bilingual Lexicon Induction (BLI), that is, using the resulting cross-lingual embeddings to build a bilingual dictionary. For that purpose, one would typically induce the translation of each source word by taking its corresponding nearest neighbor in the target language. However, it has been argued that this basic approach suffers from the hubness problem\footnote{Hubness \citep{radovanovic2010hubs,radovanovic2010existence} refers to the phenomenon of a few points being the nearest neighbors of many other points in high-dimensional spaces, which has been reported to severely affect cross-lingual embedding mappings \citep{dinu2015improving}.}, which has motivated alternative retrieval methods like inverted nearest neighbor\footnote{The original paper refers to this method as \textit{globally corrected} retrieval.} \citep{dinu2015improving}, inverted softmax \citep{smith2017offline}, and Cross-domain Similarity Local Scaling (CSLS) \citep{conneau2018word}.

In this paper, we go one step further and, rather than directly inducing the bilingual dictionary from the cross-lingual word embeddings, we use them to build an unsupervised machine translation system, and extract a bilingual dictionary from a synthetic parallel corpus generated with it. This allows us to take advantage of a strong language model and naturally extract translation equivalences through statistical word alignment. At the same time, our method can be used as a drop-in replacement of traditional retrieval techniques, as it can work with any cross-lingual word embeddings and it does not require any additional resource besides the monolingual corpus used to train them. Our experiments show the effectiveness of this alternative approach, which outperforms the previous best retrieval method by 4 accuracy points on average, establishing a new state-of-the-art in the standard MUSE dataset. As such, we conclude that, contrary to recent trend, future research in BLI should not focus exclusively on direct retrieval methods. %

\section{Proposed method}
\label{sec:method}

The input of our method is a set of cross-lingual word embeddings and the monolingual corpora used to train them. In our experiments, we use fastText embeddings \citep{bojanowski2017enriching} mapped through VecMap \citep{artetxe2018robust}, but the algorithm described next can also work with any other word embedding and cross-lingual mapping method.

The general idea of our method is to to build an unsupervised phrase-based statistical machine translation system \citep{lample2018phrase,artetxe2018usmt,artetxe2019effective}, and use it to generate a synthetic parallel corpus from which to extract a bilingual dictionary. For that purpose, we first derive phrase embeddings from the input word embeddings by taking the 400,000 most frequent bigrams and and the 400,000 most frequent trigrams in each language, and assigning them the centroid of the words they contain. Having done that, we use the resulting cross-lingual phrase embeddings to build a phrase-table as described in \citet{artetxe2018usmt}. More concretely, we extract translation candidates by taking the 100 nearest-neighbors of each source phrase, and score them with the softmax function over their cosine similarities:
\[ \phi ( \bar{f} | \bar{e} ) = \frac{ \exp \left( \cos( \bar{e}, \bar{f} ) / \tau \right)}{\sum_{\bar{f'}} \exp \left(  \cos ( \bar{e}, \bar{f'} ) / \tau \right)} \]
where the temperature $\tau$ is estimated using maximum likelihood estimation over a dictionary induced in the reverse direction. In addition to the phrase translation probabilities in both directions, we also estimate the forward and reverse lexical weightings by aligning each word in the target phrase with the one in the source phrase most likely generating it, and taking the product of their respective translation probabilities.

We then combine this phrase-table with a distortion model and a 5-gram language model estimated in the target language corpus, which results in a phrase-based machine translation system. So as to optimize the weights of the resulting model, we use the unsupervised tuning procedure proposed by \citet{artetxe2019effective}, which combines a cyclic consistency loss and a language modeling loss over a subset of 2,000 sentences from each monolingual corpora.

Having done that, we generate a synthetic parallel corpus by translating the source language monolingual corpus with the resulting machine translation system.\footnote{For efficiency purposes, we restricted the size of the synthetic parallel corpus to a maximum of 10 million sentences, and use cube-pruning for faster decoding. As such, our results could likely be improved by translating the full monolingual corpus with standard decoding.} We then word align this corpus using FastAlign \citep{dyer2013simple} with default hyperparameters and the \textit{grow-diag-final-and} symmetrization heuristic. Finally, we build a phrase-table from the word aligned corpus, and extract a bilingual dictionary from it by discarding all non-unigram entries. For words with more than one entry, we rank translation candidates according to their direct translation probability.

\begin{table*}[t]
\begin{center}
\begin{small}
  \addtolength{\tabcolsep}{-1pt}
  \begin{tabular}{lcccccccccccccc}
    \toprule
    && \multicolumn{2}{c}{en-es} && \multicolumn{2}{c}{en-fr} && \multicolumn{2}{c}{en-de} && \multicolumn{2}{c}{en-ru} && \multirow{2}{*}{avg.} \\
    \cmidrule{3-4} \cmidrule{6-7} \cmidrule{9-10} \cmidrule{12-13}
    && $\rightarrow$ & $\leftarrow$ && $\rightarrow$ & $\leftarrow$ && $\rightarrow$ & $\leftarrow$ && $\rightarrow$ & $\leftarrow$ && \\
    \midrule
    Nearest neighbor && 81.9 & 82.8 && 81.6 & 81.7 && 73.3 & 72.3 && 44.3 & 65.6 && 72.9 \\
    Inv. nearest neighbor \citep{dinu2015improving} && 80.6 & 77.6 && 81.3 & 79.0 && 69.8 & 69.7 && 43.7 & 54.1 && 69.5 \\
    Inv. softmax \citep{smith2017offline} && 81.7 & 82.7 && 81.7 & 81.7 && 73.5 & 72.3 && 44.4 & 65.5 && 72.9 \\
    CSLS \citep{conneau2018word} && 82.5 & 84.7 && 83.3 & 83.4 && 75.6 & 75.3 && 47.4 & 67.2 && 74.9 \\
    \midrule
    Proposed method && \bf 87.0 & \bf 87.9 && \bf 86.0 & \bf 86.2 && \bf 81.9 & \bf 80.2 && \bf 50.4 & \bf 71.3 && \bf 78.9 \\
    \bottomrule
  \end{tabular}
\end{small}
\end{center}
\caption{ P@1 of proposed system and previous retrieval methods, using the same cross-lingual embeddings. } %
\label{tab:results}
\end{table*}

\section{Experimental settings}
\label{sec:settings}

In order to compare our proposed method head-to-head with other BLI methods, the experimental setting needs to fix the monolingual embedding training method, as well as the cross-lingual mapping algorithm and the evaluation dictionaries. In addition, in order to avoid any advantage, our method should not see any further monolingual corpora than those used to train the monolingual embeddings. Unfortunately, existing BLI datasets distribute pre-trained word embeddings alone, but not the monolingual corpora used to train them. For that reason, we decide to use the evaluation dictionaries from the standard MUSE dataset \citep{conneau2018word} but, instead of using the pre-trained Wikipedia embeddings distributed with it, we extract monolingual corpora from Wikipedia ourselves and train our own embeddings trying to be as faithful as possible to the original settings. This allows us to compare our proposed method to previous retrieval techniques in the exact same conditions, while keeping our results as comparable as possible to previous work reporting results for the MUSE dataset.

More concretely, we use WikiExtractor\footnote{\url{https://github.com/attardi/wikiextractor}} to extract plain text from Wikipedia dumps, and pre-process the resulting corpus using standard Moses tools \citep{koehn2007moses} by applying sentence splitting, punctuation normalization, tokenization with aggressive hyphen splitting, and lowercasing. We then train word embeddings for each language using the skip-gram implementation of fastText \citep{bojanowski2017enriching} with default hyperparameters, restricting the vocabulary to the 200,000 most frequent tokens. The official embeddings in the MUSE dataset were trained using these exact same settings, so our embeddings only differ in the Wikipedia dump used to extract the training corpus and the pre-processing applied to it, which is not documented in the original dataset.

Having done that, we map these word embeddings to a cross-lingual space using the unsupervised mode in VecMap \citep{artetxe2018robust}, which builds an initial solution based on the intra-lingual similarity distribution of the embeddings and iteratively improves it through self-learning. Finally, we induce a bilingual dictionary using our proposed method and evaluate it in comparison to previous retrieval methods (standard nearest neighbor, inverted nearest neighbor, inverted softmax\footnote{Inverted softmax has a temperature hyperparameter $T$, which is typically tuned in the training dictionary. Given that we do not have any training dictionary in our fully unsupervised settings, we use a fixed temperature of $T=30$, which was also used by some previous authors \citep{lample2018phrase}. While we tried other values in our preliminary experiments, but we did not observe any significant difference.} and CSLS). Following common practice, we use precision at 1 as our evaluation measure.\footnote{We find a few out-of-vocabularies in the evaluation dictionary that are likely caused by minor pre-processing differences. In those cases, we use copying as a back-off strategy (i.e. if a given word is not found in our induced dictionary, we simply leave it unchanged). In any case, the percentage of out-of-vocabularies is always below 1\%, so this has a negligible effect in the reported results.}

\section{Results and discussion}
\label{sec:results}

Table \ref{tab:results} reports the results of our proposed system in comparison to previous retrieval methods. As it can be seen, our method obtains the best results in all language pairs and directions, with an average improvement of 6 points over nearest neighbor and 4 points over CSLS, which is the best performing previous method. These results are very consistent across all translation directions, with an absolute improvement between 2.7 and 6.3 points over CSLS. Interestingly, neither inverted nearest neighbor nor inverted softmax are able to outperform standard nearest neighbor, presumably because our cross-lingual embeddings are less sensitive to hubness thanks to the symmetric re-weighting in VecMap \citep{artetxe2018generalizing}. At the same time, CSLS obtains an absolute improvement of 2 points over nearest neighbor, only a third of what our method achieves. This suggests that, while previous retrieval methods have almost exclusively focused on addressing the hubness problem, there is a substantial margin of improvement beyond this phenomenon. %

\begin{table*}[t]
\begin{center}
\begin{small}
  \addtolength{\tabcolsep}{-1pt}
  \begin{tabular}{lcccccccccccccc}
    \toprule
    && \multicolumn{2}{c}{en-es} && \multicolumn{2}{c}{en-fr} && \multicolumn{2}{c}{en-de} && \multicolumn{2}{c}{en-ru} && \multirow{2}{*}{avg.} \\
    \cmidrule{3-4} \cmidrule{6-7} \cmidrule{9-10} \cmidrule{12-13}
    && $\rightarrow$ & $\leftarrow$ && $\rightarrow$ & $\leftarrow$ && $\rightarrow$ & $\leftarrow$ && $\rightarrow$ & $\leftarrow$ && \\
    \midrule
    \citet{conneau2018word} && 81.7 & 83.3 && 82.3 & 82.1 && 74.0 & 72.2 && 44.0 & 59.1 && 72.3 \\
    \citet{hoshen2018nonadversarial} && 82.1 & 84.1 && 82.3 & 82.9 && 74.7 & 73.0 && 47.5 & 61.8 && 73.6 \\
    \citet{grave2018unsupervised} && 82.8 & 84.1 && 82.6 & 82.9 && 75.4 & 73.3 && 43.7 & 59.1 && 73.0 \\
    \citet{alvarezmelis2018gromov} && 81.7 & 80.4 && 81.3 & 78.9 && 71.9 & 72.8 && 45.1 & 43.7 && 69.5 \\
    \citet{yang2018learning} && 79.9 & 79.3 && 78.4 & 78.9 && 71.5 & 70.3 && - & - && - \\
    \citet{mukherjee2018learning} && 84.5 & 79.2 && - & - && - & - && - & - && - \\
    \citet{alvarez2018towards} && 81.3 & 81.8 && 82.9 & 81.6 && 73.8 & 71.1 && 41.7 & 55.4 && 71.2 \\
    \citet{xu2018unsupervised} && 79.5 & 77.8 && 77.9 & 75.5 && 69.3 & 67.0 && - & - && - \\
    \midrule
    Proposed method && \bf 87.0 & \bf 87.9 && \bf 86.0 & \bf 86.2 && \bf 81.9 & \bf 80.2 && \bf 50.4 & \bf 71.3 && \bf 78.9 \\
    \bottomrule
  \end{tabular}
\end{small}
\end{center}
\caption{Results of the proposed method in comparison to previous work (P@1). All systems are fully unsupervised and use fastText embeddings trained on Wikipedia with the same hyperparameters.} %
\label{tab:sota}
\end{table*}

So as to put these numbers into perspective, Table \ref{tab:sota} compares our method to previous results reported in the literature.\footnote{Note that previous results are based on the pre-trained embeddings of the MUSE dataset, while we had to train our embeddings to have a controlled experiment (see Section \ref{sec:settings}). In any case, our embeddings are trained following the official dataset setting, using Wikipedia, the same system and hyperparameters, so our results should be roughly comparable.} As it can be seen, our proposed method obtains the best published results in all language pairs and directions, outperforming the previous state-of-the-art by a substantial margin. Note, moreover, that these previous systems mostly differ in their cross-lingual mapping algorithm and not the retrieval method, so our improvements are orthogonal.

We believe that, beyond the substantial gains in this particular task, our work has \textbf{important implications} for future research in cross-lingual word embedding mappings. While most work in this topic uses BLI as the only evaluation task, \citet{glavas2019properly} recently showed that BLI results do not always correlate well with downstream performance. In particular, they observe that some mapping methods that are specifically designed for BLI perform poorly in other tasks. Our work shows that, besides their poor performance in those tasks, these BLI-centric mapping methods might not even be the optimal approach to BLI, as our alternative method, which relies on unsupervised machine translation instead of direct retrieval over mapped embeddings, obtains substantially better results without requiring any additional resource. As such, we argue that 1) future work in cross-lingual word embeddings should consider other evaluation tasks in addition to BLI, and 2) future work in BLI should consider other alternatives in addition to direct retrieval over cross-lingual embedding mappings.

\section{Related work}
\label{sec:related}

While BLI has been previously tackled using count-based vector space models \citep{vulic2013study} and statistical decipherment \citep{ravi2011deciphering,dou2012large}, these methods have recently been superseded by cross-lingual embedding mappings, which work by aligning independently trained word embeddings in different languages. For that purpose, early methods required a training dictionary, which was used to learn a linear transformation that mapped these embeddings into a shared cross-lingual space \citep{mikolov2013exploiting,artetxe2018generalizing}. The resulting cross-lingual embeddings are then used to induce the translations of words that were missing in the training dictionary by taking their nearest neighbor in the target language.

The amount of required supervision was later reduced through self-learning methods \citep{artetxe2017learning}, and then completely eliminated through adversarial training \citep{zhang2017adversarial,conneau2018word} or more robust iterative approaches combined with initialization heuristics \citep{artetxe2018robust,hoshen2018nonadversarial}.
At the same time, several recent methods have formulated embedding mappings as an optimal transport problem \citep{zhang2017earth,grave2018unsupervised,alvarezmelis2018gromov}.

In addition to that, a large body of work has focused on addressing the hubness problem that arises when directly inducing bilingual dictionaries from cross-lingual embeddings, either through the retrieval method \citep{dinu2015improving,smith2017offline,conneau2018word} or the mapping itself \citep{lazaridou2015hubness,shigeto2015ridge,joulin2018loss}. While all these previous methods directly induce bilingual dictionaries from cross-lingually mapped embeddings, our proposed method combines them with unsupervised machine translation techniques, outperforming them all by a substantial margin.

\section{Conclusions and future work}
\label{sec:conclusions}

We propose a new approach to BLI which, instead of directly inducing bilingual dictionaries from cross-lingual embedding mappings, uses them to build an unsupervised machine translation system, which is then used to generate a synthetic parallel corpus from which to extract bilingual lexica. Our approach does not require any additional resource besides the monolingual corpora used to train the embeddings, and outperforms traditional retrieval techniques by a substantial margin. We thus conclude that, contrary to recent trend, future work in BLI should not focus exclusively in direct retrieval approaches, nor should BLI be the only evaluation task for cross-lingual embeddings. Our code is available at \url{https://github.com/artetxem/monoses}.

In the future, we would like to further improve our method by incorporating additional ideas from unsupervised machine translation such as joint refinement and neural hybridization \citep{artetxe2019effective}. In addition to that, we would like to integrate our induced dictionaries in other downstream tasks like unsupervised cross-lingual information retrieval \citep{litschko2018unsupervised}.

\section*{Acknowledgments}

This research was partially supported by the Spanish MINECO (UnsupNMT TIN2017‐91692‐EXP and DOMINO PGC2018-102041-B-I00, cofunded by EU FEDER), the BigKnowledge project (BBVA foundation grant 2018), the UPV/EHU (excellence research group), and the NVIDIA GPU grant program. Mikel Artetxe was supported by a doctoral grant from the Spanish MECD.

\bibliography{acl2019}
\bibliographystyle{acl_natbib}

\end{document}